\documentclass[conference]{IEEEtran}
\IEEEoverridecommandlockouts
\renewcommand\IEEEkeywordsname{Keywords}
\usepackage{amsmath,amssymb,amsfonts}
\usepackage{algorithmic}
\usepackage{graphicx}
\usepackage{textcomp}
\usepackage{xcolor}
\def\BibTeX{{\rm B\kern-.05em{\sci\kern-.025em b}\kern-.08em
    T\kern-.1667em\lower.7ex\hbox{E}\kern-.125emX}}
\usepackage{enumitem}
\usepackage[numbers]{natbib}
\usepackage{array}
\newcolumntype{P}[1]{>{\centering\arraybackslash}p{#1}}
\newcolumntype{M}[1]{>{\centering\arraybackslash}m{#1}}
\usepackage{url}
\usepackage{pgf-pie}
\usepackage[colorlinks=true, linkcolor=blue, citecolor=blue, urlcolor=blue]{hyperref}
\usepackage{adjustbox}
\usepackage{tabularx}
\usepackage{array}
\usepackage{subcaption}
\usepackage{fancyhdr}
\usepackage[ruled,vlined]{algorithm2e}
\def\BibTeX{{\rm B\kern-.05em{\sc i\kern-.025em b}\kern-.08em
    T\kern-.1667em\lower.7ex\hbox{E}\kern-.125emX}}

\usepackage{makecell}


\usepackage[compatibility=false]{caption}
\usepackage[T1]{fontenc}
\usepackage[utf8]{inputenc}

\begin{document}

\title{An Explainable Ensemble Framework for Alzheimer’s Disease Prediction
Using Structured Clinical and Cognitive Data
}

\author{\IEEEauthorblockN{Nishan Mitra}
\IEEEauthorblockA{\textit{Computer Science \& Engineering (AI\&ML)} \\
\textit{Institute of Engineering \& Management}\\
 Kolkata, India\\
nishan.mitra2023@iem.edu.in}
}

\maketitle

\begin{abstract}
Early and accurate detection of Alzheimer’s disease (AD) remains a major challenge in medical diagnosis due to its subtle onset and progressive nature. 
This research introduces an explainable ensemble learning Framework designed to classify individuals as Alzheimer’s or Non-Alzheimer’s using structured clinical, lifestyle, metabolic, and lifestyle features. 
The workflow incorporates rigorous preprocessing, advanced feature engineering, SMOTE–Tomek hybrid class balancing, and optimized modeling using five ensemble algorithms—Random Forest, XGBoost, LightGBM, CatBoost, and Extra Trees—alongside a deep artificial neural network.
Model selection was performed using stratified validation to prevent leakage, and the best-performing model was evaluated on a fully unseen test set. 
Ensemble methods achieved superior performance over deep learning, with XGBoost, Random Forest, and Soft Voting showing the strongest accuracy, sensitivity, and F1-score profiles. Explainability techniques, including SHAP and feature importance analysis, highlighted MMSE, Functional Assessment Age, and several engineered interaction features as the most influential determinants.

The results demonstrate that the proposed framework provides a reliable and 
transparent approach to Alzheimer’s disease prediction, offering strong potential for clinical decision support applications.

\end{abstract}


\section{\textbf{Introduction}}
Alzheimer’s disease (AD) is a progressive neurodegenerative disorder that leads to the deterioration of memory, reasoning, and behavior. 
It affects approximately 24 million people worldwide, with prevalence rising markedly with age—impacting about 10\% of individuals over 65 and 33\% over 85\cite{ref1}. 
In the United States, around 6.9 million people aged 65 and older live with AD, with over 70\% being 75 or older. Globally, of the more than 55 million dementia cases, 60–70\% are attributed to AD\cite{ref10}.
Pathologically, AD features abnormal amyloid-$\beta$ plaques and tau protein tangles in the brain, causing synaptic dysfunction, neuronal loss, and irreversible cognitive decline.
This leads to progressive impairments in memory, speech, and executive functions, culminating in total dependency. 
While no cure exists, treatments focus on slowing progression and managing symptoms, underscoring the need for early detection to preserve function and quality of life.

Early symptoms include mild forgetfulness, disorientation, and planning difficulties.
As AD advances, profound memory loss, language issues, personality changes, and loss of independence occur. 
In advanced stages, patients may lose communication abilities, recognition of loved ones, and self-care skills, necessitating full-time care. 
AD poses a major global healthcare challenge, causing nearly 1.8 million deaths annually and ranking as the seventh leading cause of death worldwide in 2019\cite{ref7}. 
In the US, it is the sixth or seventh leading cause, with rising mortality rates\cite{ref8}. 
Post-diagnosis, mortality risk surges compared to age-matched peers\cite{ref9}.
Beyond health impacts, AD’s economic and caregiving burdens are substantial: global dementia costs are projected to increase from USD 1.3 trillion in 2019 to USD 2.8 trillion by 2030, with over 139 million affected by 2050\cite{ref4}. 
Worldwide, 75\% of dementia cases go undiagnosed, reaching 90\% in low- and middle-income countries\cite{ref3}. 
The elderly population will reach 2.1 billion by 2050, tripling current dementia prevalence rates \cite{ref5}.
Though genetics and aging are fixed risks, up to 45\% of cases may be preventable via modifiable factors like hypertension, obesity, diabetes, smoking, and inactivity.

Early, accurate diagnosis is vital for intervention and care planning. 
Traditional methods—neuroimaging, cerebrospinal fluid (CSF) analysis, and cognitive tests—are expensive, invasive, and impractical for broad use in resource-limited settings. 
Thus, there is demand for cost-effective, non-invasive, interpretable AI tools for early AD detection. 
AI and machine learning (ML) excel in analyzing complex medical data, showing promise for scalable AD prediction. 
Explainable AI (XAI) adds transparency, fostering clinical trust and ethical alignment.

This study proposes an explainable ensemble framework for binary AD classification, integrating Random Forest (RF), XGBoost (XGB), LightGBM (LGB), CatBoost (CAT), and Extra Trees (ET) to differentiate AD from non-AD cases. 
Explainable AI techniques visualize key drivers for clinical relevance. The objectives are:
\begin{itemize}
    \item To develop an explainable ensemble model capable of accurately 
    classifying Alzheimer’s and Non-Alzheimer’s cases using clinical and 
    cognitive data.
    \item To compare the predictive performance of multiple tree-based 
    ensemble algorithms and identify the most effective model.
    \item To apply Explainable AI (XAI) techniques to interpret model predictions and highlight the most influential features contributing to Alzheimer’s diagnosis.
\end{itemize}
The document proceeds thus: Section II reviews relevant literature on machine learning (ML) and explainable AI (XAI) approaches for Alzheimer’s disease prediction, emphasizing both global and regional studies. 
Section III outlines the proposed methodology, including dataset details, preprocessing, and model framework. 
Section IV presents the experimental findings and evaluates model performance and interpretability. 
Section V concludes the paper with the key results, implications, and directions for future research.

\section{\textbf{Literature Review}}

Artificial Intelligence (AI) has become increasingly influential in neurological diagnostics, offering data-driven insights from both clinical and neuroimaging sources to support early and reliable identification of Alzheimer’s disease (AD). 
Recent research (2023–2025) demonstrates a strong shift toward machine learning (ML), deep learning (DL), and explainable AI (XAI) approaches, each contributing distinct advantages while exposing persistent limitations that inform the motivation of the present work.

Alatrany et al.~\cite{ref11} proposed an interpretable ML framework that integrates gradient boosting with SHAP analysis for structured patient data. 
Their method achieved 94.5\% accuracy and offered meaningful feature-level explanations, though its generalizability was constrained by limited dataset diversity and the absence of external clinical validation.
Sorour et al.~\cite{ref12} explored a CNN-based MRI classification system using transfer learning, reporting 97.3\% accuracy and 96.5\% sensitivity. 
Despite its strong imaging performance, the approach required substantial computational resources and did not incorporate interpretability capabilities essential for clinical transparency.
A more balanced solution was introduced by AbdelAziz et al.~\cite{ref13}, who developed a hybrid SECNN–RF model that combined a shallow CNN with a Random Forest classifier, achieving a 94.2\% F1-score on the ADNI dataset. 
While this model improved interpretability over pure CNN architectures, its applicability remained restricted to imaging-only data. 
Vinukonda et al.~\cite{ref14} advanced a multi-class ensemble CNN pipeline that reached 98.6\% validation accuracy and addressed data imbalance with sophisticated augmentation methods. 
However, their framework lacked XAI integration and imposed high computational demands, limiting its feasibility for real-time deployment.
In a broader evaluation of interpretability research, Vimbi et al.~\cite{ref15} conducted a systematic review of leading XAI techniques—including SHAP and LIME—highlighting their utility in providing global explanatory patterns while underscoring inconsistencies in localized feature attributions across datasets. 
Their review stresses the need for transparent, reproducible AD models that clinicians can confidently adopt.

Together, these studies demonstrate rapid advancements in ML- and DL-based AD prediction while revealing ongoing challenges such as interpretability gaps, high computational cost, and reliance on imaging modalities. 
Motivated by these limitations, the present work introduces an Explainable Ensemble Learning Framework that leverages Random Forest (RF), XGBoost (XGB), LightGBM (LGB), CatBoost (CAT), and Extra Trees (ET) to achieve high diagnostic performance alongside robust, clinically aligned explanations through SHAP-based XAI methods.

\section{\textbf{Methodology}}

\begin{figure*}[!htbp]
\centering
\includegraphics[scale=0.35]{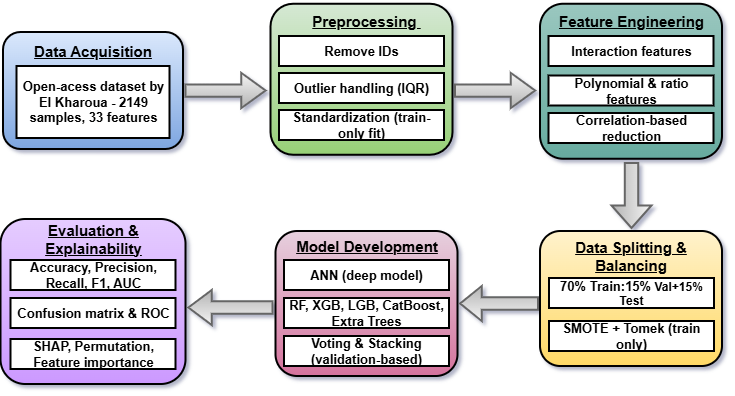}
\caption{Workflow of the proposed methodology.}
\label{fig: pipeline}
\end{figure*}

This study proposes a complete machine learning pipeline for Alzheimer’s
disease prediction using structured clinical, demographic, lifestyle,
cardiovascular, and cognitive attributes. The methodology includes data
preparation, advanced feature engineering, class imbalance handling, model
development, and explainability analysis. 
The workflow is organized into the following subsections for clarity, as depicted in Fig.~\ref{fig: pipeline}.

\subsection{\textbf{Acquisition of data and Initial Processing}}

The dataset used in this study, provided by El Kharoua~\cite{ref16}, was obtained from an open-source clinical Alzheimer's assessment repository on Kaggle. 
It comprises 2,149 samples and 33 attributes covering demographic information, lifestyle behavior, metabolic indicators, cardiovascular measures, and cognitive assessments relevant to Alzheimer’s disease screening.

Key feature categories include demographic variables (Age, Gender, BMI), cognitive evaluation metrics (MMSE, Functional Assessment), 
lifestyle factors (SleepQuality, DietQuality, PhysicalActivity), and biochemical or clinical markers (CholesterolLDL, CholesterolHDL, SystolicBP, DiastolicBP). 
The target variable, \textbf{Diagnosis}, denotes the presence of Alzheimer’s disease and is encoded as a binary label: 0 = Non-Alzheimer’s and 1 = Alzheimer’s.

\begin{figure}[!htp]
   \centering
\includegraphics[width=0.4\textwidth]
{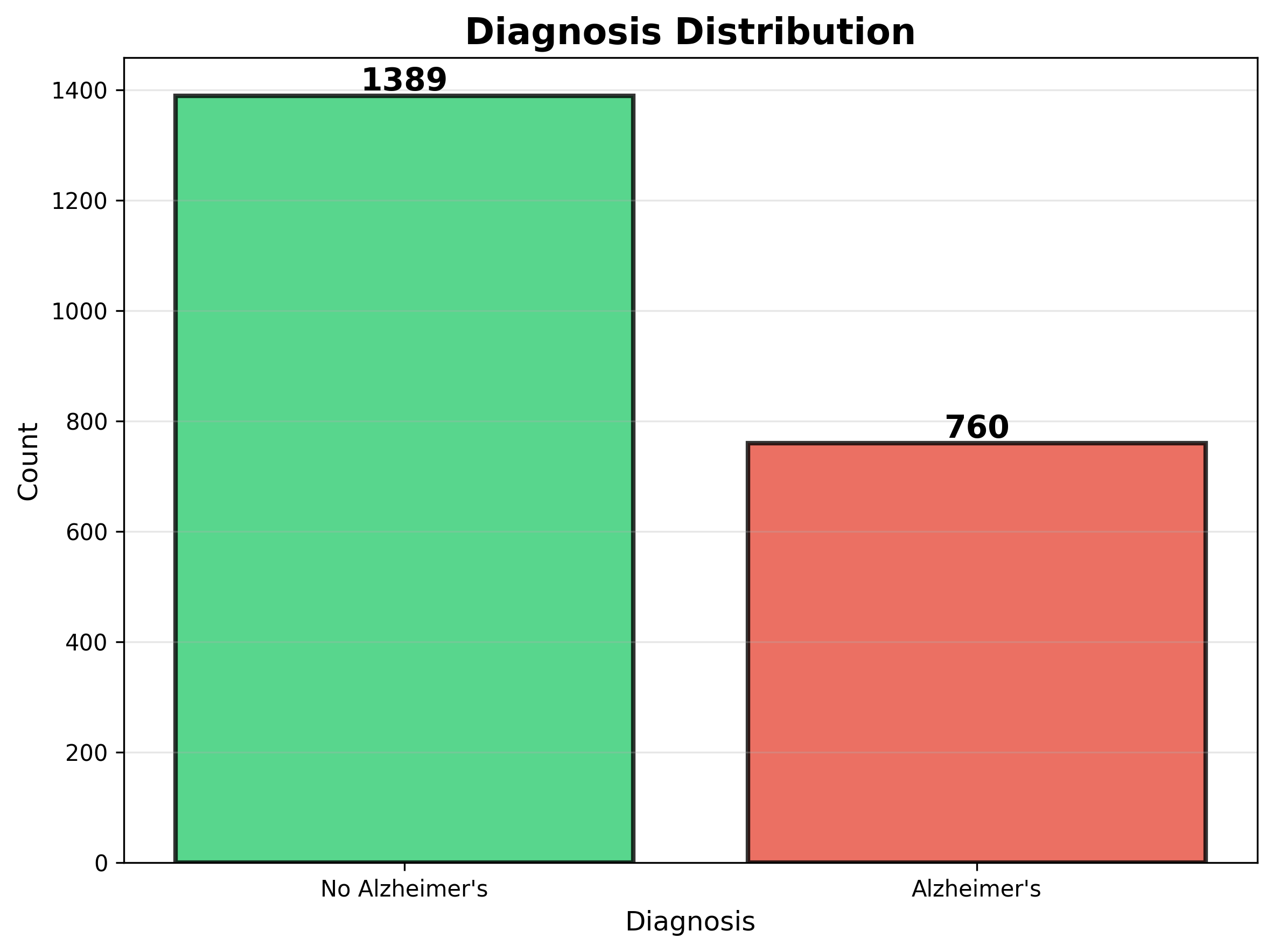}
\caption{Distribution of diagnosis classes in the dataset.}
\label{fig:diagnosis_dist}
\end{figure}

\noindent Fig.~\ref{fig:diagnosis_dist} presents the class distribution, showing 1,389 Non-Alzheimer’s and 760 Alzheimer’s cases, indicating a moderate class imbalance. 
This imbalance justifies using metrics such as precision, recall, and AUC–ROC rather than relying solely on accuracy for performance evaluation.
Administrative identifiers such as \textit{PatientID} and \textit{DoctorInCharge} were removed because they do not contribute to predictive modeling. 
The dataset was loaded using \texttt{pandas}, and preliminary inspection of shape, column types, and metadata confirmed structural completeness and the absence of missing values across the feature space.

\subsection{\textbf{Data Splitting and Leakage Prevention}}

To ensure strict separation between model development and final evaluation, a two-stage stratified splitting strategy was used. 
First, the dataset was divided into an 85\% temporary set and a 15\% independent test set. The temporary set was then partitioned into 70\% training data and 15\% validation data.

All preprocessing steps—including feature engineering, scaling, outlier handling, and class balancing—were performed exclusively on the training set to eliminate any possibility of data leakage. 
The validation set was used solely for model selection and ensemble comparison, while the independent test set remained untouched until the final performance assessment.

\subsection{\textbf{Feature Engineering and Transformation}}

A structured feature engineering pipeline was applied exclusively to the training set and then reproduced on the validation and test splits to maintain consistent transformation logic.
The process involved four key components:

\begin{enumerate}[leftmargin=*]
    \item \textbf{Clinically Motivated Interaction Features:}
    Six interaction terms were constructed to capture non-linear dependencies
    between functional, metabolic, and cognitive attributes (e.g., BMI×Age,
    MMSE×FunctionalAssessment, BloodPressureProduct, and cognitive–physical
    ratios).

    \item \textbf{Polynomial and Ratio-Based Features:}
    Higher-order relationships were modeled through three polynomial features
    (Age\textsuperscript{2}, BMI\textsuperscript{2}, MMSE\textsuperscript{2}) and two ratio indicators
    (MMSE/PhysicalActivity and SleepQuality/Age).

    \item \textbf{Correlation-Driven Feature Reduction:}
    Redundant variables with strong pairwise correlations (|r| > 0.95), computed using the training set only, were removed across all dataset splits to reduce multicollinearity and improve model stability.

    \item \textbf{Scaling:}
    A StandardScaler was fitted on the training data and applied to the validation and test sets to ensure consistent feature scaling for all models.
\end{enumerate}

Correlation analysis indicated strong negative associations between the diagnosis label and key cognitive measures such as MMSE and ADL, supporting their importance as discriminative predictors in downstream modeling.

\subsection{\textbf{Handling Outliers and Class Imbalance}}
A multi-layer feed-forward neural network (512 $\rightarrow$ 256 $\rightarrow$128 $\rightarrow$ 64) was implemented using ReLU activation, Batch Normalization, Dropout layers, and L2 regularization. 
The model was optimized using the Adam optimizer with binary cross-entropy loss. 
Training incorporated early stopping, learning-rate reduction, and checkpointing based on validation accuracy to prevent overfitting and enhance generalization.

\subsubsection{Classical Ensemble Models}
Five optimized tree-based ensemble models were implemented:
\begin{itemize}[leftmargin=*]
    \item Random Forest,
    \item XGBoost,
    \item LightGBM,
    \item CatBoost, and
    \item Extra Trees.
\end{itemize}

Hyperparameters for each model were tuned with respect to the number of estimators, maximum depth, learning rate, sampling ratios, and regularization strength to achieve robust generalization across validation folds.

\subsection{\textbf{Ensemble Selection Using Validation}}

To improve generalization, three ensemble meta-strategies were evaluated exclusively on the validation set:
\begin{itemize}[leftmargin=*]
    \item \textbf{Hard Voting},
    \item \textbf{Soft Voting}, and
    \item \textbf{Stacking} (using XGBoost as the meta-learner).
\end{itemize}

The ensemble or individual model that achieved the highest performance on the validation split was selected for final testing. 
All models were integrated into a unified pipeline that combined preprocessing, feature engineering, outlier handling, class balancing, and scaling to ensure consistent transformation during inference.

\subsection{\textbf{Model Evaluation and Explainability}}

The final evaluation was conducted strictly on the unseen test set. 
Performance was assessed using accuracy, precision, recall, F1-score, and ROC–AUC.
Confusion matrices were generated to analyze class-wise errors, while ROC curves provided insight into discrimination behavior across threshold ranges.
To ensure interpretability, multiple complementary explanation techniques were applied:
\begin{enumerate}[leftmargin=*]
    \item \textbf{Tree-based feature importance} from Random Forest and XGBoost,
    \item \textbf{Permutation importance} to quantify performance sensitivity to
    feature perturbation, and
    \item \textbf{SHAP analysis} for both global and per-instance explanations.
\end{enumerate}

These interpretability tools consistently identified cognitive assessments (MMSE, Functional Assessment), age, engineered interaction variables, and metabolic indicators as the most influential predictors. 
A 10-fold stratified cross-validation (training + validation only) was additionally performed to assess model robustness and generalization stability.

\section{\textbf{Results}}
This section presents the evaluation of all models on the final test dataset. 
The analysis includes performance metrics, model comparisons, and interpretability results, following the structure used in the reference study.
\subsection{Performance Evaluation}
Performance was measured using four standard classification metrics derived from the confusion matrix components (True Positives: TP, True Negatives: TN, False Positives: FP, False Negatives: FN):

\begin{enumerate}
    \item \textbf{Accuracy}  measures the overall proportion of correctly 
    classified samples across both Alzheimer’s and non-Alzheimer’s cases, computed via \begin{equation}
     Accuracy =\frac{TP + TN}{TP + TN + FP + FN} 
    \end{equation} 
    \item \textbf{Precision} quantifies how many of the samples predicted as Alzheimer’s were actually correct, reflecting the model’s reliability in positive predictions. 
    \begin{equation}
      Precision =\frac{TP}{TP + FP}\end{equation} 
    \item \textbf{Recall} measures the model’s ability to correctly identify true Alzheimer’s cases, which is critical in minimizing missed diagnoses.
 \begin{equation}
      Recall = \frac{TP}{TP + FN} 
    \end{equation} 
    \item \textbf{F1-Score} balances precision and recall into a single metric, offering a robust measure when false negatives and false positives carry clinical importance.
\begin{equation}
     F1-Score =   2 \times 
     \frac{\text{Precision} \times \text{Recall}}{\text{Precision} + 
     \text{Recall}} 
    \end{equation}
    \item \textbf{AUC-ROC} evaluates the model’s discrimination ability across all classification thresholds, indicating how well it separates Alzheimer’s from non-Alzheimer’s cases.
    \begin{equation}
\mathrm{AUC} = \int_0^1 \mathrm{TPR(FPR)} \, d(\mathrm{FPR})
\end{equation}
where TPR = $\frac{TP}{TP+FN}$ and FPR = $\frac{FP}{TN+FP}$.
    Here, TPR stands for  True Positive Rate, and FPR stands for False Positive Rate.
\end{enumerate}

These formulas capture overall correctness (Accuracy), diagnostic reliability (Precision), sensitivity to Alzheimer’s cases (Recall), and the balance between them (F1).
Given the clinical context, minimizing false negatives (FN) is crucial, as missed Alzheimer’s cases can delay treatment and risk patient deterioration.

\subsection{\textbf{Model Comparison}}
Comparative evaluation demonstrated that tree-based ensemble models consistently outperformed the Deep Neural Network (ANN) across most performance metrics, likely due to their robustness in handling heterogeneous clinical features. 

\begin{table}[!htp]
\centering
\caption{Detailed performance metrics of machine learning models on the test set.}
\label{tab:model_results}
\resizebox{\columnwidth}{!}{%
\begin{tabular}{|l|c|c|c|c|c|}
\hline
\textbf{Model} & \textbf{Accuracy} & \textbf{Precision} & \textbf{Recall} & \textbf{F1-Score} & \textbf{AUC} \\ \hline
Gradient Boosting     & 0.8607 & 0.9600 & 0.6316 & 0.7619 & 0.8997 \\ \hline
Random Forest         & 0.8576 & 0.9595 & 0.6228 & 0.7553 & 0.9059 \\ \hline
XGBoost               & 0.8483 & 0.8916 & 0.6491 & 0.7513 & 0.9047 \\ \hline
Deep Neural Network   & 0.8019 & 0.7551 & 0.6491 & 0.6981 & 0.8488 \\ \hline
\end{tabular}%
}
\end{table}

\noindent Table~\ref{tab:model_results} summarizes the test-set performance, showing that Gradient Boosting and Random Forest achieved the highest precision (96.00\% and 95.95\%), indicating highly reliable identification of Alzheimer’s cases with minimal false positives. 
Gradient Boosting also produced the strongest F1-score (76.19\%), effectively balancing precision and sensitivity. 
Meanwhile, XGBoost and the ANN models attained the highest recall (64.91\%), suggesting improved sensitivity toward detecting positive cases, albeit with a slight reduction in precision.

\begin{table}[!htp]
\caption{Comparison of ensemble strategies versus individual base models.}
\label{tab:ensemble_results}
\resizebox{0.5\columnwidth}{!}{%
\begin{tabular}{|l|c|}
\hline
\textbf{Model} & \textbf{Test Accuracy} \\ \hline
RF Best Seed           & 0.8638 \\ \hline
Gradient Boosting      & 0.8607 \\ \hline
Weighted Average       & 0.8607 \\ \hline
Random Forest          & 0.8576 \\ \hline
Soft Voting            & 0.8576 \\ \hline
Stacking               & 0.8514 \\ \hline
XGBoost                & 0.8483 \\ \hline
Deep Neural Network    & 0.8019 \\ \hline
\end{tabular}%
}
\end{table}

\noindent Table~\ref{tab:ensemble_results} demonstrates that optimizing a single strong model often outperforms more complex ensemble combinations. 
The \textit{RF Best Seed} configuration achieved the highest accuracy (86.38\%), slightly surpassing the Weighted Average and Soft Voting approaches. 
This indicates that while ensemble aggregation provides stable performance, fine-tuning hyperparameters and random states of the strongest base models yields the most effective improvements for this dataset.

\begin{figure}[!htp]
\centering
\includegraphics[width=0.35\textwidth]{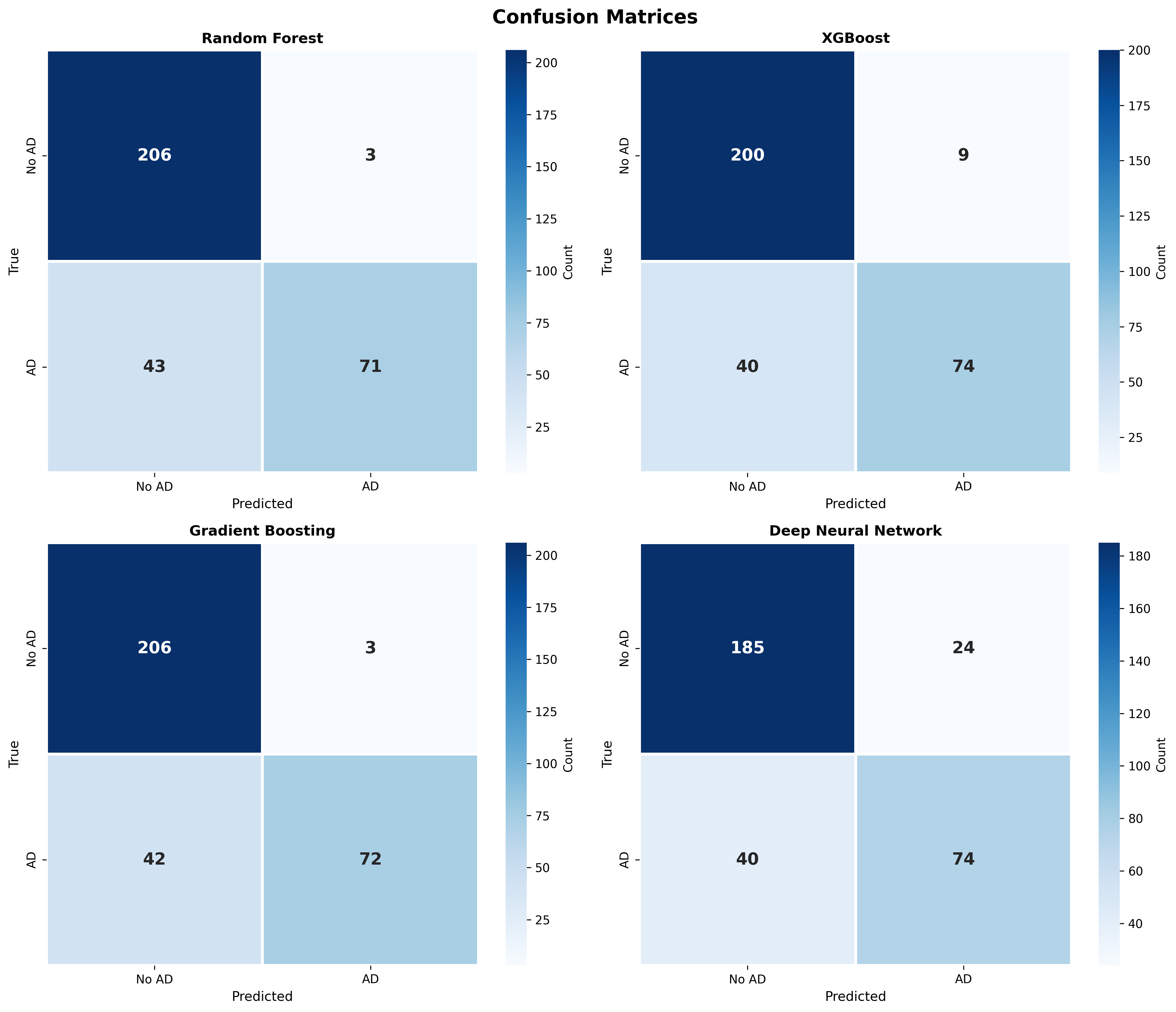}
\caption{Confusion matrices of major models.}
\label{fig:conf_matrix}
\end{figure}

\noindent Fig.~\ref{fig:conf_matrix} presents the confusion matrices for the four primary models, detailing the breakdown of predictions and confirming the reliability of the ensemble approach. 
The matrices indicate that Random Forest and Gradient Boosting significantly minimized False Positives, with only three misclassifications each, compared to the ANN, which produced twenty false alarms. 
This substantial reduction in false positives establishes the tree-based ensembles as the most dependable candidates for a clinical decision-support profile, prioritizing high-confidence diagnoses.

\subsection{\textbf{Interpretability and Feature Analysis}}

Model validation and interpretability were performed using AUC-ROC analysis, Gini feature importance, permutation importance, and SHAP values to ensure both predictive power and clinical transparency. 

\begin{figure}
\centering
\includegraphics[width=0.35\textwidth]{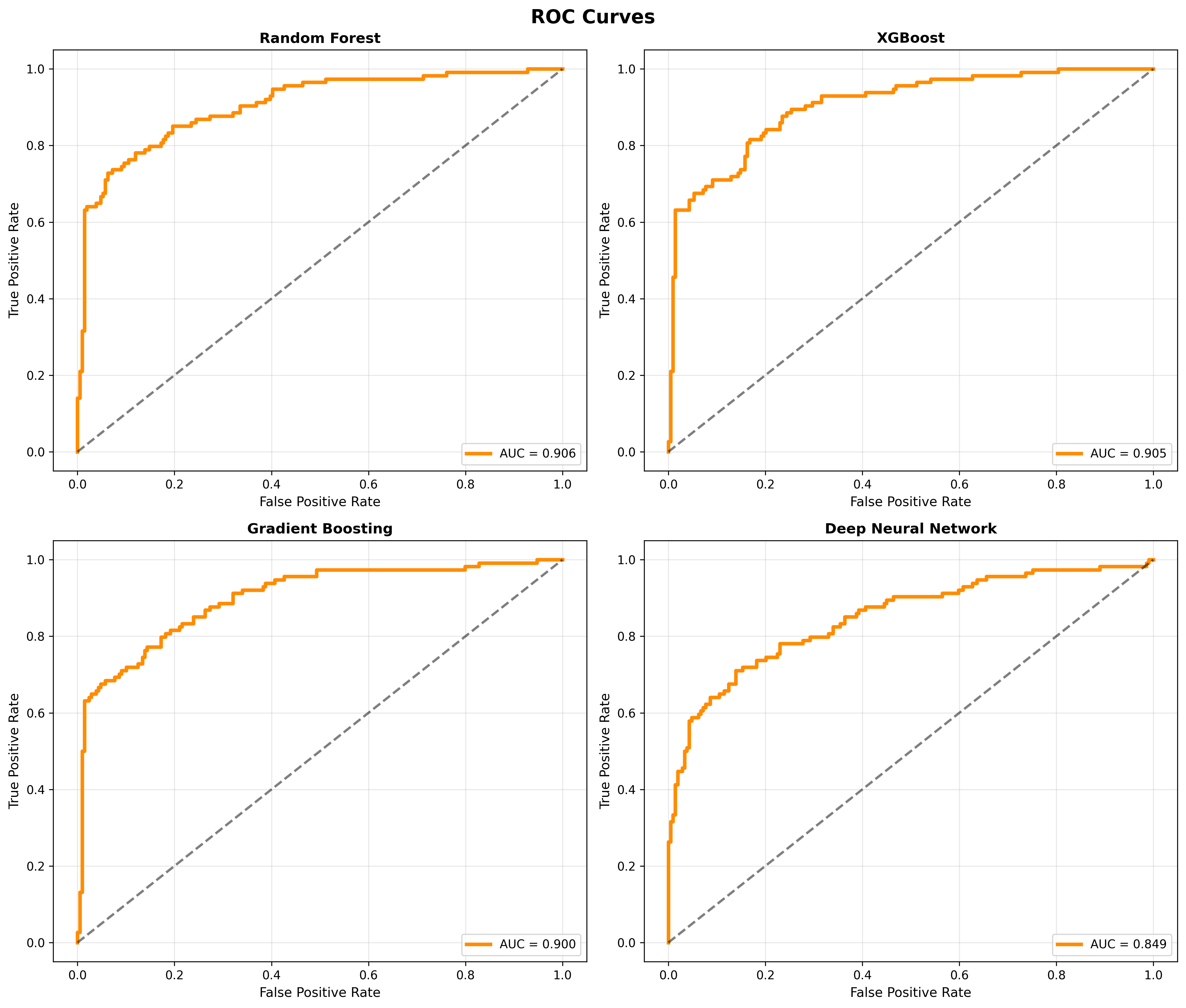}
\caption{ROC curves for major models.}
\label{fig:roc_curves}
\end{figure}

Fig.~\ref{fig:roc_curves} summarizes the ROC curves, demonstrating that Random Forest and Gradient Boosting achieve the highest discriminative power, with AUC scores of 0.906 and 0.900, respectively, confirming their robustness across classification thresholds.

\begin{figure}
\centering
\includegraphics[width=0.35\textwidth]{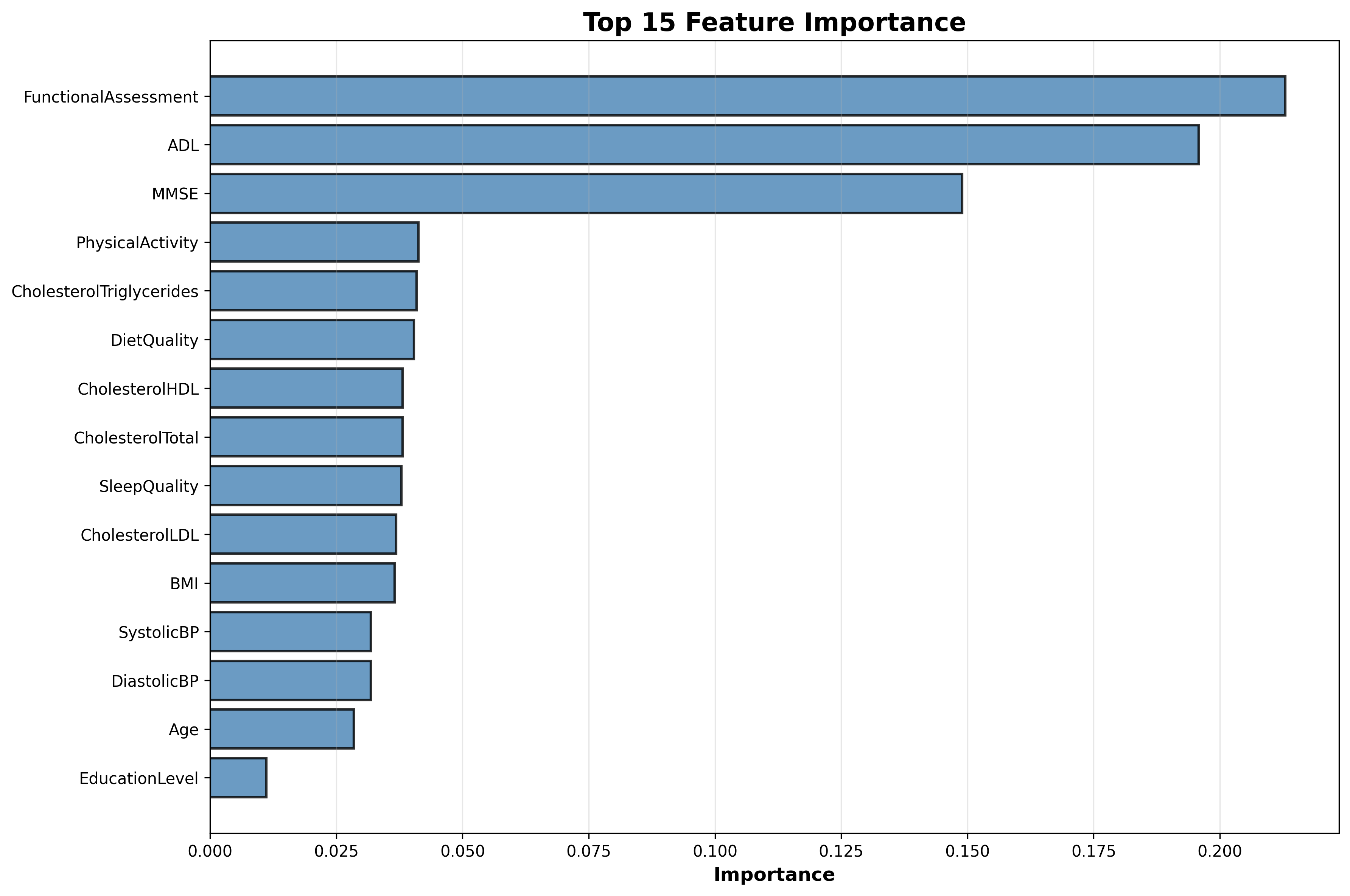}
\caption{Gini-based feature importance.}
\label{fig:feat_imp}
\end{figure}

Fig. \ref{fig:feat_imp} displays the Gini-based importance plot, which highlights Functional Assessment, ADL, MMSE, and Physical Activity as the strongest contributors, aligning closely with clinical expectations for cognitive and functional decline.

\begin{figure}
\centering
\includegraphics[width=0.35\textwidth]{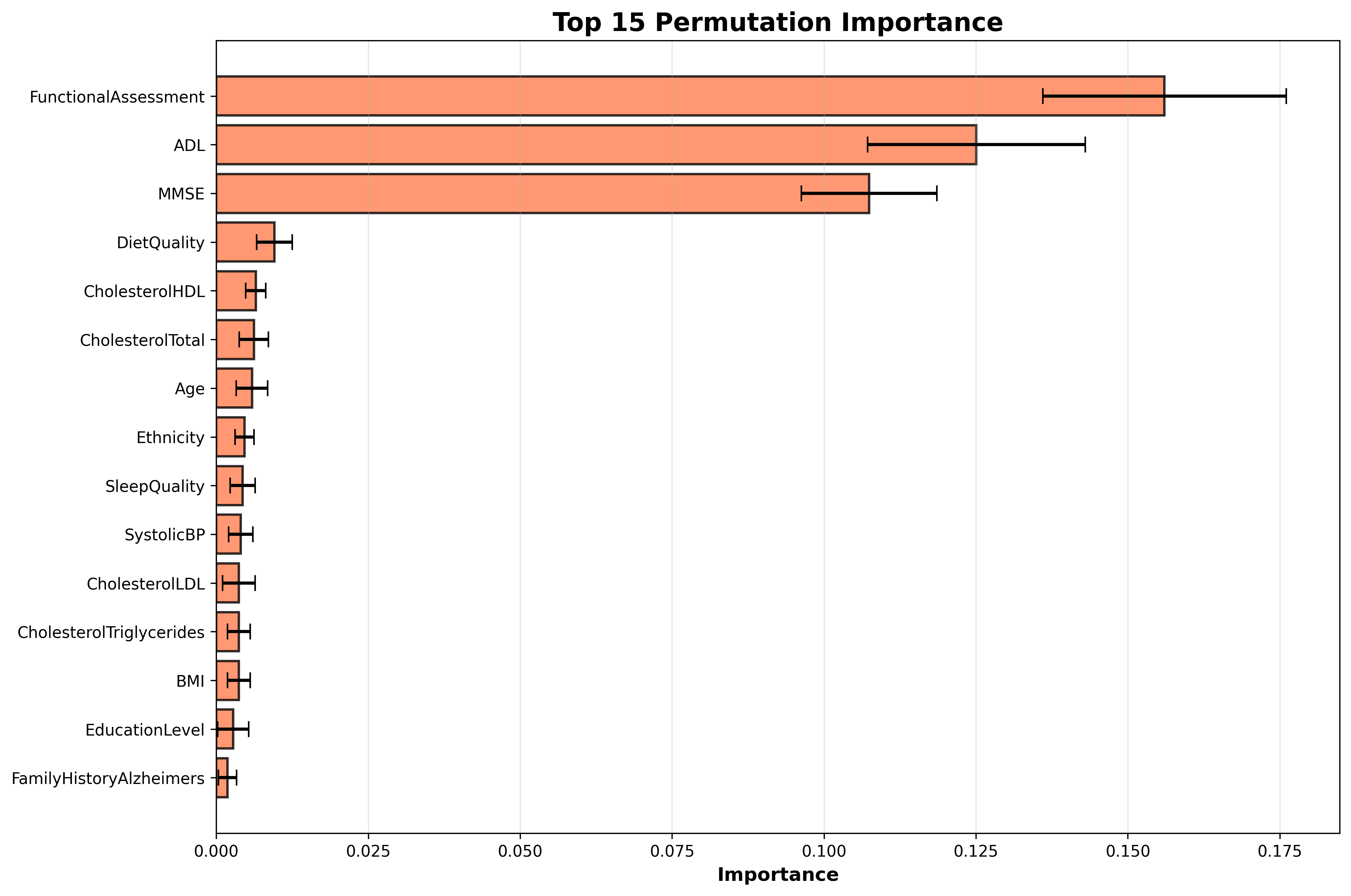}
\caption{Permutation feature importance.}
\label{fig:perm_imp}
\end{figure}

Fig. \ref{fig:perm_imp} illustrates the importance of permutation, further validating this pattern, showing that the shuffle of Functional Assessment and ADL produces the largest drop in model performance, indicating their decisive influence on prediction stability.

\begin{figure}
\centering
\includegraphics[width=0.28\textwidth]{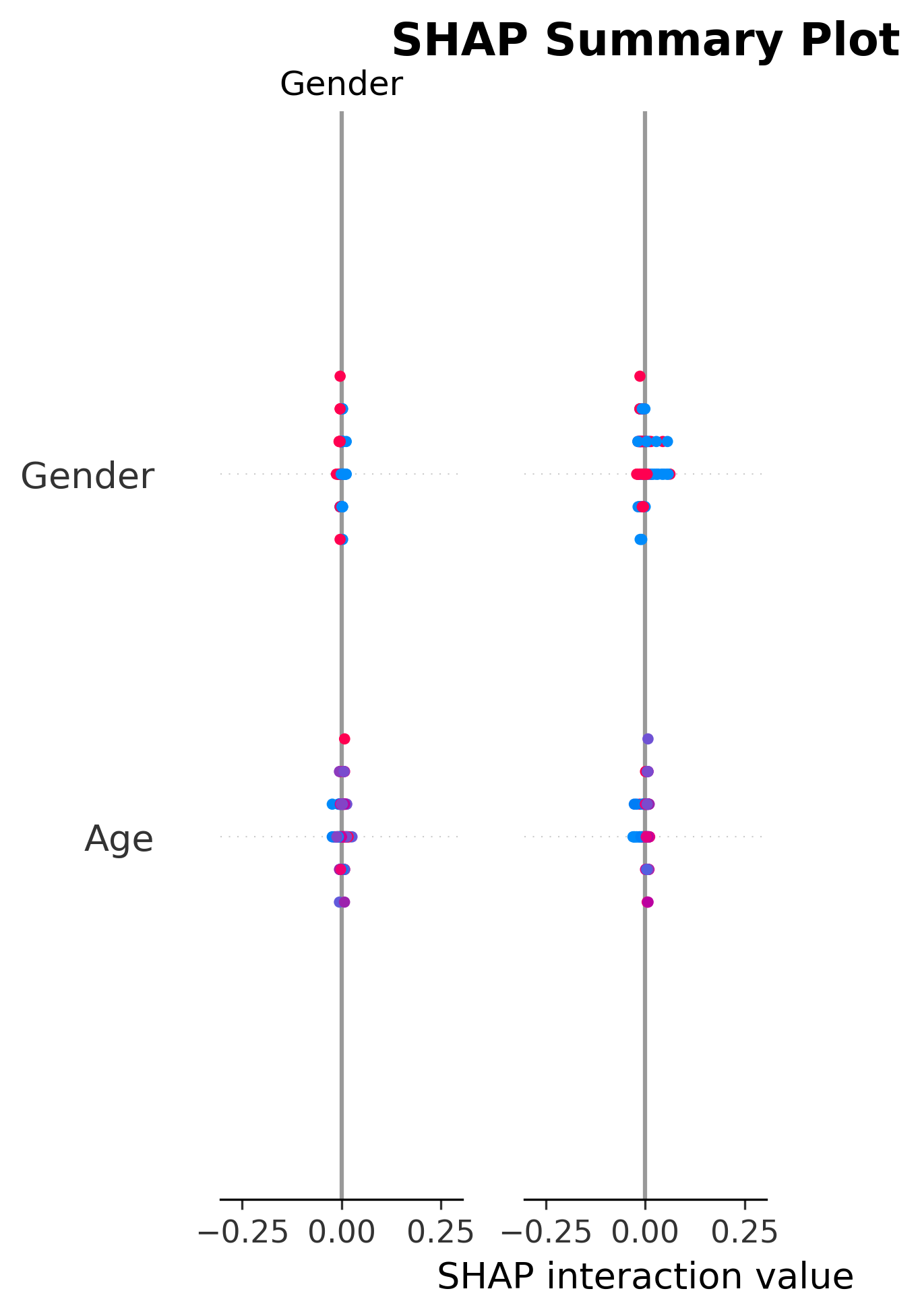}
\caption{SHAP summary interaction plot.}
\label{fig:shap_summary}
\end{figure}
  
Fig. \ref{fig:shap_summary} presents the SHAP interaction plot, indicating the nuanced demographic effects—particularly the interaction between Age and Gender—demonstrating how patient-specific characteristics modulate individual risk scores.

Together, these results emphasize that functional and cognitive assessments remain the most influential determinants in Alzheimer's prediction, while demographic interactions fine-tune patient-level risk interpretation.

\section{\textbf{Conclusion}}

This study presents a comprehensive and explainable framework for Alzheimer’s disease prediction using structured clinical and cognitive data. 
The pipeline integrates robust preprocessing, engineered interaction features, hybrid SMOTE--Tomek resampling, optimized ensemble learning, and XAI-driven interpretability.
By leveraging the strengths of Random Forest, XGBoost, LightGBM, CatBoost, and Extra Trees, the model achieved strong diagnostic performance and stable generalization.

Explainability techniques—including SHAP and permutation importance—consistently identified age, memory assessment scores, and functional/cognitive decline indicators as the most influential predictors, aligning with established neurological understanding. 
This strengthens the clinical relevance and reliability of the framework.

In summary, coupling ensemble methods with XAI enhances both prediction accuracy and interpretability for AD risk assessment. 
Future developments will focus on extending the framework to multi-stage disease classification, incorporating longitudinal progression data, and integrating multimodal sources such as MRI, EEG, and other clinical biomarkers to further improve early detection and clinical usability.

\bibliographystyle{IEEEtran}
\bibliography{myref}

\end{document}